%% file: Kenya_vaccine.tex
\documentclass[sigconf]{acmart}

\usepackage{times}
\usepackage{amsmath, amsfonts}
\usepackage{multirow}
\usepackage{graphicx, subfigure}
\usepackage{multirow, booktabs}
\usepackage{caption}
\usepackage{balance}
\usepackage{bm}
\usepackage{booktabs}      
\usepackage{float}         
\usepackage{array}  
\usepackage{placeins}
\usepackage{graphicx}
\usepackage{longtable} 

\setcopyright{acmcopyright}
\acmConference[Green Lab 2020/2021]{Green Lab 2020/2021 - Vrije Universiteit Amsterdam}{September--October, 2020}{Amsterdam, The Netherlands}
\acmBooktitle{Green Lab 2020/2021 - Vrije Universiteit Amsterdam, September--October, 2020, Amsterdam, The Netherlands}

\title{Using Synthetic Data for Machine Learning-based Childhood Vaccination Prediction in Narok, Kenya}

\author{Jimmy Bach}
\orcid{0009-0008-9573-2955}
\affiliation{
 \institution{William \& Mary}
 \city{Williamsburg, VA}
 \country{United States}
}
\affiliation{%
  \institution{Ignite Global Health Research Lab, William \& Mary Global Research Institute}
  \city{Williamsburg}
  \state{VA}
  \country{USA}
}

\author{Yang Li, M.S.}
\orcid{0009-0008-3409-0936}
\affiliation{%
  \institution{William \& Mary Department }
  \city{Williamsburg}
  \state{VA}
  \country{USA}
}
\affiliation{%
  \institution{Data-Driven Decision Intelligence Lab (D3i)}
  \city{Williamsburg}
  \state{VA}
  \country{USA}
}

\author{Yaqi Liu, B.S.}
\affiliation{
 \institution{William \& Mary}
 \city{Williamsburg, VA}
 \country{United States}
}

\author{John Sankok}
\affiliation{
  \institution{Community Health Partners}
  \city{Narok}
  \country{Kenya}
}

\author{Rose Kimani}
\affiliation{
  \institution{Community Health Partners}
  \city{Narok}
  \country{Kenya}
}

\author{Carrie B. Dolan, PhD}
\orcid{0000-0002-4445-7075}

\affiliation{%
  \institution{William \& Mary Health Sciences}
  \city{Williamsburg}
  \state{VA}
  \country{USA}
}

\affiliation{%
  \institution{Ignite Global Health Research Lab, William \& Mary Global Research Institute}
  \city{Williamsburg}
  \state{VA}
  \country{USA}
}
\author{Julius N. Odhiambo, PhD}
\orcid{0000-0002-9817-1849}

\affiliation{%
  \institution{William \& Mary Health Sciences}
  \city{Williamsburg}
  \state{VA}
  \country{USA}
}
\affiliation{%
  \institution{Ignite Global Health Research Lab, William \& Mary Global Research Institute}
  \city{Williamsburg}
  \state{VA}
  \country{USA}
}
\author{Haipeng Chen, PhD}
\orcid{0000-0003-0572-8888}
\affiliation{%
  \institution{William \& Mary Department of Data Science}
  \city{Williamsburg}
  \state{VA}
  \country{USA}
}
\affiliation{%
  \institution{Data-Driven Decision Intelligence Lab (D3i)}
  \city{Williamsburg}
  \state{VA}
  \country{USA}
}

\begin{document}

\begin{abstract}
\textbf{Background:} 
Limited data utilization in low-resource settings poses a major barrier to the vaccine delivery ecosystem, undermining efforts to achieve equitable immunization coverage. In nomadic populations, where reliable data are hard to obtain, individuals face an increased risk of missing crucial vaccination doses as children. One such population is the Maasai in Narok County, Kenya, where the absence of high-volume, high-quality data hampers accurate coverage estimates, impedes efficient resource allocation, and weakens the ability to design and deliver timely interventions. Additionally, data privacy concerns are heightened in groups with limited sensitive data. 
\textbf{Objectives:} First, we aim to identify children at risk of missing key vaccines across a large population to provide timely, evidence-based interventions that support increased vaccination coverage. Second, we aim to better protect the privacy of sensitive health data within a vulnerable population.
\textbf{Methods:} We digitized 8 years of child vaccination records from the MOH 510 registry (n=6,913) and applied machine learning models (i.e., Logistic Regression and XGBoost) to identify children at risk. Additionally, we utilize a novel approach to tabular diffusion-based synthetic data generation (``TabSyn'') to protect patient privacy within the models.
\textbf{Results:} Our findings show that classification techniques can reliably and successfully predict children at risk of missing a vaccine, with recall, precision, and F1-scores exceeding 90\% for some vaccines modeled. Additionally, training these models with synthetic data rather than real data—preserving the privacy of individuals within the original dataset—does not lead to a loss in predictive performance. 
\textbf{Conclusion:} These results support the use of synthetic data implementation in health informatics strategies for clinics with limited digital infrastructure, enabling privacy-preserving, scalable forecasting for childhood immunization coverage.
\end{abstract}
\maketitle

\section{Introduction}


Inequities in childhood vaccination coverage remain a persistent challenge. The third dose of the diphtheria, tetanus, and pertussis vaccine (DTP3)—a standard indicator of routine immunization performance—achieved 85\% global coverage in 2024, yet approximately 20 million children remained unvaccinated or under-vaccinated. Of these, 14.3 million were “zero-dose” children, an increase from 12.9 million in 2019, underscoring the ongoing inequities in vaccine access \citep{manyanga2025decade, UNICEF2024}. These gaps are most pronounced in low- and middle-income countries (LMICs), where geographic isolation, socioeconomic barriers, and political instability hinder sustainable progress toward universal immunization and health equity. In sub-Saharan Africa, countries such as Rwanda, Kenya, and Ghana have integrated vaccination into their primary healthcare systems and expanded outreach through community health workers and mobile units. Nevertheless, disparities persist, particularly among nomadic populations whose migratory lifestyles impede consistent access to vaccination services and increase vulnerability to outbreaks.

These persistent gaps highlight the need for data-driven strategies to help health systems identify and prioritize children at the greatest risk of missing vaccinations \cite{chen2022using,xiao2022sequential}. Timely identification of such populations could allow more efficient targeting of limited resources, particularly in low-capacity environments. This is crucial for meeting the goals of global health commissions and organizations \cite{who_ia2030_2020, gavi_6_strategy_narrative_2024, who_sdg3_gap_2019}. Setbacks such as conflict, climate change, and misinformation have made reaching these goals increasingly unlikely, making an equity-centered approach supported by high-quality data a requirement to get back on track \cite{haeuser2025global}. While these frameworks outline ambitious goals, progress also depends on empirical innovations tested in local contexts. 

Predictive modeling offers a promising opportunity to address inequities by enabling healthcare leaders to proactively identify children at risk of missing vaccine doses. Unlike traditional surveillance systems, predictive models can leverage historical and demographic data to guide targeted interventions, optimize limited resources, and improve health equity \cite{ou2021active}. However, in many low-resource settings, several barriers constrain implementation of predictive modeling. Infrastructure limitations—including inconsistent electricity, poor internet connectivity, and limited equipment—reduce the capacity to build, maintain, and interpret digital health systems. Moreover, incomplete or unreliable data often prevent health professionals from training robust models. 

In Narok County, a largely rural district in southwest Kenya, these issues remain prominent during data collection. Record-keeping systems in the area are often paper-based registries, which raises multiple concerns. First, these registries frequently contain missing data, negatively impacting a clinician's ability to provide the correct resources to their patients. Additionally, the time required to collect, analyze, and use this data for enhanced service increases, making timely intervention difficult \cite{orwa2022management}.

These challenges are compounded by privacy and cybersecurity concerns, as weak data protection measures increase the risk of patient data breaches. \cite{masinde2025big}. While Kenya has developed legislation to address these concerns through the 2019 Data Protection Act, questions remain about its effectiveness in implementation and enforcement, including within the healthcare sector.  For children in small, identifiable populations, such as certain sub-communities within the Maasai, data re-identification remains a concern, and the disclosure of health and vaccination information could lead to misuse. Together, these concerns may limit the ability to share and pool data in machine learning applications.

To overcome these barriers, synthetic data generation has emerged as a promising solution. Over the past decade, advances in machine learning—such as variational autoencoders (VAEs) \cite{kingma2022autoencodingvariationalbayes}, generative adversarial networks (GANs) \cite{goodfellow2014generative}, and most recently diffusion models—have enabled the probabilistic generation of new, realistic data samples that mimic the statistical properties of original datasets \cite{ho2020denoising, li2025population}. In healthcare, researchers use synthetic data to augment or replace real data for model training, helping preserve patient privacy while addressing issues of data scarcity and fragmentation.

Our work builds on this emerging field through collaboration with Community Health Partners (CHP), the largest community healthcare provider in a nomadic setting in Kenya. CHP currently relies on paper-based vaccination registries, such as the Ministry of Health’s MOH 510 form, which presents multiple challenges for digital transformation. Cleaning and digitizing records takes significant time, making real-data-based prediction difficult or infeasible. For community health providers like CHP, who operate in resource-constrained environments, synthetic data offers a cost-efficient, scalable alternative for developing predictive insights into vaccination coverage.

Building on previous efforts to digitize community-level health data in Kenya, we have the following research questions:
\begin{itemize}
    \item \textbf{RQ1:} How can we accurately predict childhood vaccination outcomes from Narok County using machine learning?  
    \item \textbf{RQ2:} Can we preserve the privacy of vulnerable populations by training models on synthetic data similar to original data without degrading model performance? 
\end{itemize}
To address \textbf{RQ1}, we utilize Logistic Regression and Extreme Gradient Boosting (XGBoost) classifiers for individual childhood vaccination outcomes associated with 17 different vaccines. To answer \textbf{RQ2}, we first utilize Mixed-Type Tabular Data Synthesis with Score-Based Diffusion in Latent Space (TabSyn) \cite{zhang2024mixedtype}, a SOTA latent diffusion-based tabular data generation method, which can generate high-fidelity synthetic data based on our collected data. Then, we retrain models using synthetic data. 
Results show varying levels of success in applying machine learning models to the original CHP registry data. The models achieved high predictive accuracy for early-stage vaccines such as BCG—often exceeding 90\% precision, recall, and F1-scores—but struggled to achieve such accuracy in vaccines administered later due to more mixed levels of coverage in the dataset. The results also show that synthetic data can effectively replace real patient data without significant loss of model performance. Metrics were consistently similar between models trained on real and synthetic data. The generated data maintained over 95\% fidelity to the original data structure. The overall results confirm the utility of synthetic data in low-resource environments.

To summarize, this study shows that we can use machine learning and synthetic data methods to develop a scalable, privacy-preserving pathway for community health providers to strengthen immunization systems in low-resourced communities. To the best of our knowledge, this is the first study to examine childhood vaccination coverage in a low-resource nomadic setting, while preserving data privacy by using tabular diffusion methods.
As a pilot study, the results show the potential to extend predictive capacity to low-resourced communities, enabling targeted interventions that advance health equity and move countries closer to achieving Universal Health Coverage and Sustainable Development Goal 3.8 (Vaccines for All) \cite{forslund2024strengthening}.

\section{Related Works}



\paragraph{Machine Learning for vaccine modeling.} Machine learning (ML) methods have increasingly been applied to healthcare domains (e.g, predict childhood vaccination uptake and micronutrient deficiency prediction \cite{bondi2022micronutrient, bondi2023predicting}). Applications of vaccination have been increasingly studied in the aftermath of the COVID-19 pandemic, during which Artificial Intelligence (AI) and ML were frequently used in pandemic response \cite{gawande2025role} and for predicting COVID vaccine uptake \cite{cheong2021predictive, dodoo2024using}. These methods have also been applied to childhood vaccination, with many studies focusing on LMIC contexts where gaps in immunization remain large. These applications often rely on nationally representative survey datasets, such as the Demographic and Health Surveys (DHS), to model determinants of incomplete immunization. Across diverse settings—including Bangladesh \cite{hasan2021associating}, Zimbabwe \cite{mbunge2025leveraging}, Ethiopia \cite{demsash2023machine}, and Tanzania \cite{kalegele2025determinants}—tree-based and ensemble methods such as Decision Trees, Random Forests, and Extreme Gradient Boosting (XGBoost) have been the most commonly used approaches. These models consistently identify sociodemographic, geographic, and maternal health factors as significant predictors of vaccine uptake. While these studies demonstrate the feasibility and importance of ML-based immunization prediction, they are predominantly conducted using national-level survey data, which may obscure important sub-national heterogeneity and limit applicability to specific local contexts.

While the potential for AI and machine learning to advance health equity through clinical support in childhood vaccination is immense, a litany of limitations often prevents that potential from being realized. First, much of the current research in AI within LMIC countries depends on national-level survey datasets due to gaps in data availability at subnational levels. While the reliability of the data used in these surveys is strong due to rigid statistical techniques carried out by organizations like USAID with the DHS, the nature of the data makes it unclear whether the study results are generalizable to local communities, particularly those that do not visit structured health systems consistently and instead utilize the services of community health clinics.  In current solutions, many models in LMICs are trained on High Income Country (HIC) data, which is more accessible but may not generalize well to LMIC settings \cite{lopez2022challenges}. Finally, limited attention has been given to privacy-preserving techniques that enable modeling without exposing sensitive health information in patient populations. This study takes key steps to fill these gaps, ensuring that the data collected is contextually relevant, locally representative, and privacy-preserving when applying tree-based ML models to LMIC vaccination coverage. 

\paragraph{Synthetic Tabular Data Generation}

Data scarcity and privacy constraints remain significant hurdles in healthcare research. Synthetic data generation offers a viable solution to this bottleneck, producing datasets that preserve the statistical properties of original distributions \cite{li2025population}. Early approaches employed Generative Adversarial Networks (GANs) like ITS-GAN \cite{chen2019faketables}, to generate synthetic relational tabular data. GAN-based methods such as medGAN \cite{armanious2020medgan} and medBGAN \cite{baowaly2019synthesizing}, or Variational Autoencoders (VAEs), such as EVA \cite{biswal2021eva} have also been used to model Electronic Health Records (EHR). Diffusion Models (DMs) \cite{ho2020denoising,song2021scorebased} have recently emerged as a superior generative paradigm, yielding high-fidelity data for images \cite{ho2020denoising,rombach2022high}, time series \cite{li2025population,yuan2024diffusionts}, and tabular formats \cite{zhang2024mixedtype,kotelnikov2023tabddpm}. Notably, models like ScoEHR \cite{naseer2023scoehr} demonstrate generation quality that outperforms previous methods. Despite initial success, the application of data synthesis to vaccination coverage and demand modeling remains unexplored. This research bridges that gap by leveraging generative models for vaccine coverage, directly supporting the equity-centered objectives of Global Health 2050 and the Immunization Agenda 2030 to improve vaccination rates and reduce premature mortality.

\section{Methodology}
In this paper, we studied childhood vaccination coverage in a low-resource community while preserving data privacy. To address \textbf{RQ1}, we developed machine learning classifiers using Logistic Regression and XGBoost models \cite{chen2016xgboost}. To address \textbf{RQ2}, we generated synthetic vaccination data samples using the TabSyn \cite{zhang2024mixedtype} framework and compared them with real CHP registry data using statistical similarity metrics and pairwise correlation analysis. After generating the synthetic data, we retrained and evaluated machine learning models under three conditions: using real data, synthetic data, and combinations of both, and compared predictive performance across these conditions. The following subsections describe the registry data, preprocessing, and model development in detail.

\subsection{CHP Dataset Collection}
Data used for this study were collected from Community Health Partners (CHP), a faith-based non-profit community health clinic collective in Narok, Kenya. The dataset was sourced from paper copies of MOH 510, the Ministry of Health's national immunization registry for children. Registry data were collected on site at each of the four community health clinics operated by CHP in Ewaso-Ngiro, Talek, Aitong, and Mara Rianta for 9,718 patients from September 2016 to June 2024. All children recorded in the registry during this time period were included in the collection process. Data from this 8-year window were collected and digitized during June and July of 2023 and 2024. The outcome of interest was whether an individual received a vaccination at any point during the study period. 

\subsubsection{Data Preprocessing Steps}

\begin{figure}
        \centering
        \includegraphics[width=\linewidth]{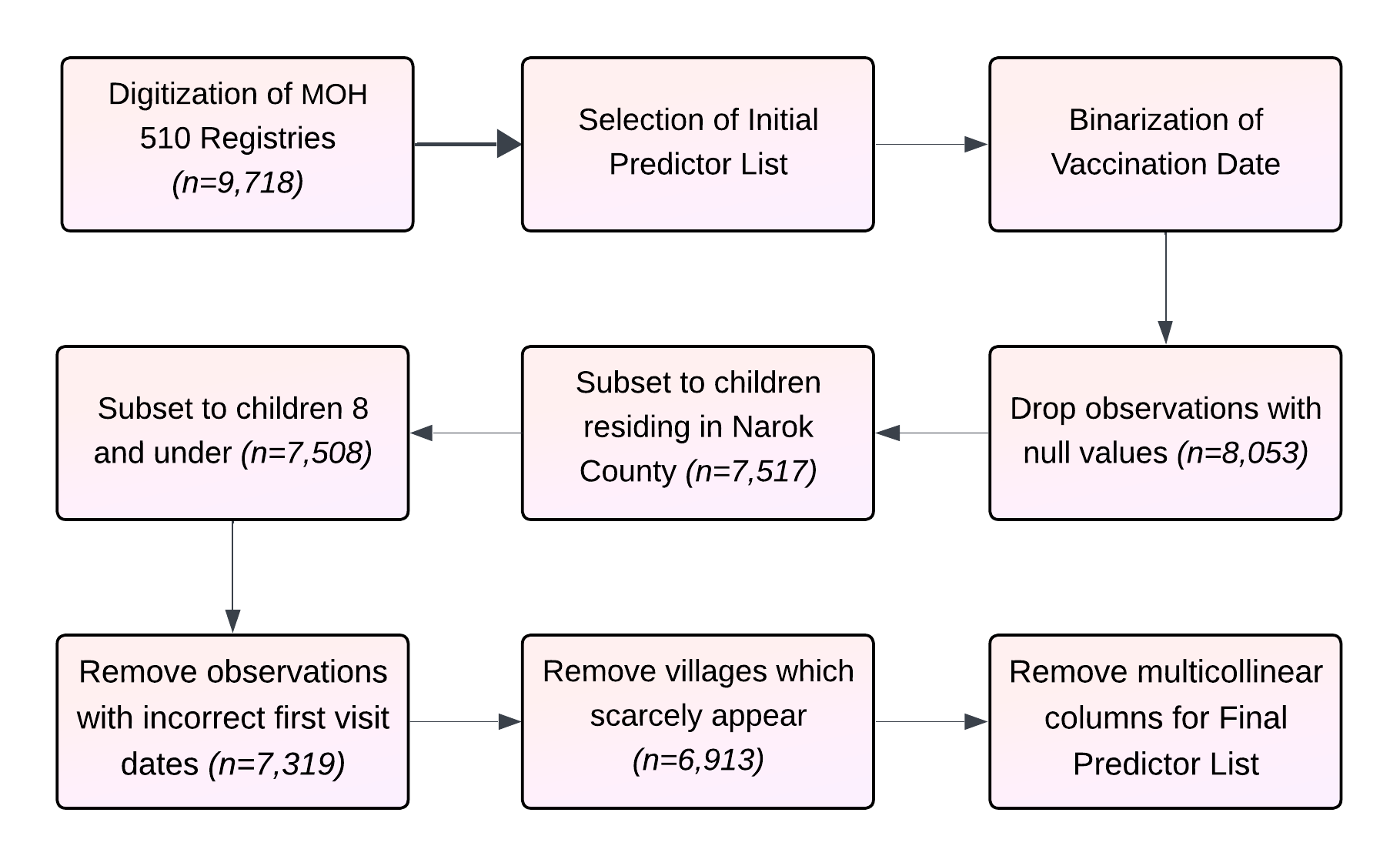}
    \caption{Data Preprocessing Steps}
     \label{fig:data_preprocessing}
    \end{figure}

Following the methodology of Phillips et al. \cite{phillips2017determinants} and leveraging expert domain knowledge, we refined our dataset to include only the most relevant predictors.
While some home villages and all clinic locations featured existing geospatial data, other variables contained unknown distributions. To maintain data integrity, we removed observations with missing and inconsistent feature values from the dataset, resulting in a final analytical sample of 6,913 patients (down from 9,718). The outcome variable was derived from recorded vaccination dates, with missing entries signifying non-vaccination. These preprocessing stages are detailed by data type below and illustrated in Figure \ref{fig:data_preprocessing}.

\paragraph{1. Categorical Data}
Categorical features from our original dataset included the child's gender, home village, and clinic name. One major preprocessing step was applied to these features. Using the location data associated with each village, it was observed that some villages had latitude and longitude coordinates that extended far beyond Narok County, the area CHP is set up to serve. Since the objective of the models was to increase and predict coverage \textit{in Narok County}, patients with home villages outside the county boundaries were removed, reducing the sample size to 7,517 patients. After this, we removed the multicollinear latitude and longitude predictors (as these are perfectly correlated with the village predictor).

\paragraph{2. Numeric Data}
Numeric features in our dataset included the child's age and the first visit day for vaccination. Each of these variables was subsetted to avoid unreliable observations. Since the data was collected from September 2016 to June 2024, we wanted to ensure that the children in the dataset were not born far before the window of observation. The age predictor represented the child's age at the end of the time of data collection (06/06/2024), so values over the age of 8 were excluded, removing 9 more patients from the dataset. With regards to the first visit day, there were numerous children with dates of first visit to a clinic that were prior to their listed date of birth. Due to concerns about the reliability of these observations, they were also removed from the dataset, reducing the dataset further to 7,319 observations.

\paragraph{3. Outcome Data}
After the initial predictors were chosen, we defined and transformed our outcome variables. Since the registry contained dates of vaccination for those who received a certain vaccine (and missing values for those who did not), the vaccine date columns were binarized such that those with a date received a value of 1 and those without received a value of 0. This was done for all 17 vaccines within the registry.

\paragraph{4. Additional Filtering for Categorical Stability}
To make the generation of synthetic data possible, each value for categorical features needed to appear in both training and testing splits. This was not an issue for any of the predictors besides the home village predictor, where some villages appeared as little as one time throughout the registry. To ensure that this would not be an issue when splitting the data, we conservatively removed any observations that had a home village that appeared fewer than 5 times in the dataset.

\paragraph{Feature Scaling}
Data was scaled using Sci-kit Learn's StandardScaler class, normalizing features to have a mean value of 0 and a standard deviation of one. Parameters were fit to the training data and later applied to the test data so as to avoid data information leakage.  

 \begin{table}[ht]
\centering
\input{table2.tex}
\caption{Summary Statistics of the Study Population}
\label{tab:summary_stats}
\end{table}

\begin{figure}
    \centering
    \includegraphics[width=0.95\linewidth]{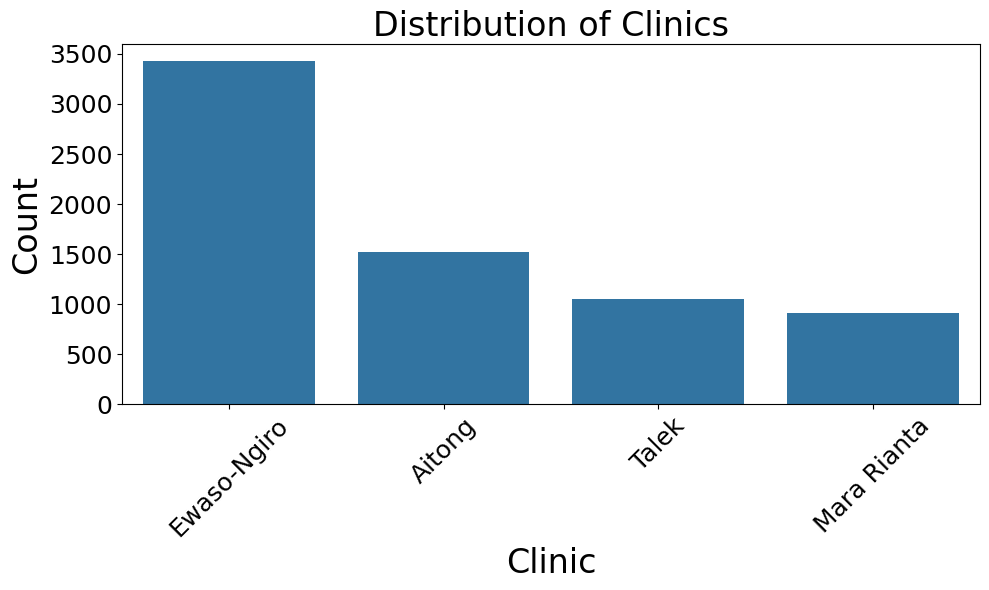}
    \caption{Number of Individuals Within Each Clinic Registry}
    \label{fig:clinic_counts}
\end{figure}

\begin{figure}
    \centering
    \includegraphics[width=0.95\linewidth]{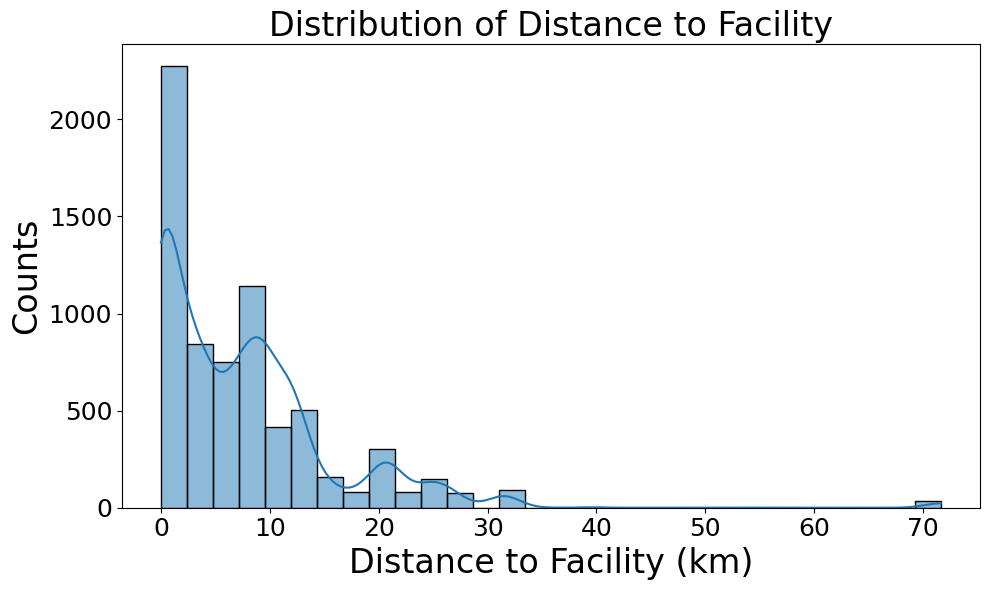}
    \caption{Distances Traveled by Individuals to the Nearest Health Facility}
    \label{fig:Distance}
\end{figure}

\paragraph{Resulting Dataset}
After preprocessing and data filtering steps, the final dataset contained 6,913 patients. To provide an overview of the dataset, Table \ref{tab:summary_stats}  displays key summary statistics, while Figure \ref{fig:clinic_counts} and Figure \ref{fig:Distance} illustrate the distribution of patients by clinic in the dataset and the distances they traveled to receive vaccinations.

Nearly half of the data were collected from the Ewaso-Ngiro clinic, the primary clinic in the region, with the remaining records in the dataset split relatively evenly among the Aitong, Talek, and Mara Rianta clinics. Most individuals were within 10 kilometers of the clinic they visited, while a substantial subset had to travel farther. The table and figures highlight the population's contextual and geographic heterogeneity in Narok County. 

In addition to contextual differences, Table \ref{tab:outcomes} illustrates the variation in vaccination outcomes for different vaccines under study. Individual vaccination coverage was high in earlier stages of the immunization schedule, such as BCG (95.6\%), but considerably decreased for later vaccines, with coverage rates as low as 13.7\% for the MR2 vaccine. Adherence to recommended childhood vaccination schedules appears to weaken as children age further, leading to coverage gaps for late-stage vaccines.

\begin{table}[h]
\centering
\input{table3.tex}
\caption{Outcome Distribution for Different Vaccines}
\label{tab:outcomes}
\end{table}

\subsection{Predictive Models}
In preparing for \textbf{RQ1}, we applied two types of machine learning models to predict vaccination uptake: Logistic Regression and XGBoost. These models were selected for their complementary strengths—Logistic Regression provides a simple and interpretable approach to binary classification, while XGBoost offers a powerful ensemble learning method capable of handling complex, non-linear relationships in the data. Logistic Regression models were built using the default implementations of the LogisticRegression class in Scikit-learn \cite{pedregosa2011scikit}, while XGBoost models \cite{chen2016xgboost} were built using the default implementations of the XGBClassifier from the XGBoost library. We were motivated to leave the models in their default state rather than tuning the hyperparameters because it simplified the process of directly comparing the success of the model when training it on different training sets. 

To conduct our analysis, we employed a train-test split approach. Specifically, 80\% of the registry data as well as different samples of synthetic data were allocated for training the models, while the remaining 20\% of registry data was reserved for testing. This method ensured that the models were trained on a majority of the data while maintaining an independent dataset to evaluate their performance.

The primary objective of \textbf{RQ2} was to investigate how the incorporation of synthetic data affects the success of predicting coverage and whether it can effectively be used to preserve the privacy of a vulnerable population. Secondarily, we wanted to observe whether the augmentation of synthetic data to the real CHP registry data could improve predictive performance. Thus, for each model, we implemented three types of training sets: one that used only real data, one that used only synthetic data, and one that combined both real and synthetic data. There were 5 different variants for each of the training sets that included synthetic data, as synthetic observations were sampled 5 separate times. For each of these sets, we also tested whether the Synthetic Minority Oversampling Technique (SMOTE) would improve the model's performance by training a model once with SMOTE and once without SMOTE, as many of the vaccines had severe class imbalances. This is implemented using the SMOTE Imbalanced Learn package. The training set's minority class was oversampled to equal the number of observations in the training set's majority class. 
\subsection{Synthetic Data}
As previously mentioned, there were inherent difficulties and limitations in data collection that provided the basis for synthetic data generation. First, paper record-keeping proved difficult to interpret. Data cleaning and preparation during digitization were imperfect and time-consuming tasks. More importantly, by using synthetic data, our dataset allowed for increased patient privacy. Within a vulnerable population with easily re-identifiable personal data, the motivation for using synthetically generated data is enhanced to ensure adequate protection of the individuals under study.

To develop synthetic samples for \textbf{RQ2}, we employed TabSyn \cite{zhang2024mixedtype}, a state-of-the-art latent diffusion model for tabular data generation. TabSyn consists of a variational autoencoder (VAE) and a score-based diffusion model, which we explain in detail in the following sections. We next describe the two components of the TabSyn architecture: the variational autoencoder (VAE) and the latent diffusion process.

\subsubsection{VAEs}
The use of VAEs was motivated by the desire for the data to be presented in a continuous embedding space. Our dataset contained both numeric and categorical variables, meaning that the distribution of some of our features is not continuous but rather discrete. Since a standard diffusion process operates under the assumption of a continuous input space, using a VAE for a continuous latent representation allows these assumptions to be met without losing information. TabSyn's VAE architecture consists of a tokenizer, a Transformer encoder, a Transformer decoder, and a detokenizer. The tokenizer pre-processes numerical columns via a quantile transformer and categorical columns via one-hot encoding, and then a linear transformation is applied to numeric columns, and an embedding lookup table is designed for the categorical columns, such that 
\begin{equation}\label{eqn:tokenizer}
\begin{split} 
    e^{num}_i  = x^{num}_i\cdot w^{num}_i + b_i^{num}, \; \;  e^{cat}_i  =  {x}^{oh}_i \cdot W^{cat}_i + b^{cat}_i,
\end{split}
\end{equation}
where $x^{num}_i, x^{oh}_i$ are the numeric and one-hot encoded columns, respectively, $w^{num}_i, b^{num}_i, b^{cat}_i \in \mathbb{R}^{1 \times d}$, $W^{cat}_i \in \mathbb{R}^{C_i \times d}$ are learnable parameters of the tokenizer, $e^{num}_i, e^{cat}_i\in \mathbb{R}^{1 \times d}$.

After being tokenized, the embeddings are passed into the Transformer encoder. The encoder module consists of a $\mu$ encoder and a $log \; \sigma$ encoder which respectively represent the mean and standard deviations of the probability distribution for the latent variable. Using the values from the matrices outputted by these two encoders, latent variables are created via \textit{reparameterization}: $\hat{z}=\mu + \epsilon\cdot\sigma$, where $\epsilon=\mathcal{N}(0,I)$. The latent embeddings then get passed through the decoder to obtain reconstructed tokens, which then pass through the detokenizer to acquire the reconstructed columns. 

Using the detokenized features ($\hat{x}$), the $\beta$-VAE loss function shown below (Eqn. \eqref{eqn:loss}) is used to update the VAE network:
\begin{equation}\label{eqn:loss}
\begin{split} 
\mathcal{L}=\ell_{recon}(x,\hat{x})+\beta\ell_{KL}(\mu,\sigma),
\end{split}
\end{equation}
where $\ell_{recon},\ell_{KL}$ are respectively the reconstruction loss and KL-divergence loss.

\subsubsection{Latent Diffusion Models}
A latent diffusion model consists of two main steps. First, the model applies an autoencoder to compress the data into a latent space. Then it diffuses the latent representation by iteratively adding noise, then denoises this transformed observation by learning the noise schedule. The forward (Eq. \ref{eqn:forward}) and reverse (Eq. \ref{eqn:reverse}) diffusion processes of Tabsyn are listed below.
\begin{align}
    {z}_t & =  {z}_0 + \sigma(t) \bm{\varepsilon}, \; \bm{\varepsilon} \sim \mathcal{N}(\bm{0}, {I}),   \label{eqn:forward}\\
    {\rm d}{z}_t & =  -2\dot{\sigma}(t) \sigma(t) \nabla_{{z}_t} \log p({z}_t) {\rm d}t + \sqrt{2\dot{\sigma}(t) \sigma(t)} {\rm d} \bm{\omega}_t,   \label{eqn:reverse}
\end{align}
where ${z}_0 = {z}$ is the initial embedding from the encoder, ${z}_t$ is the diffused embedding at time $t$, and $\sigma(t)$ is the noise level. In the reverse process, $\nabla_{{z}_t} \log p_t({z}_t)$ is the score function of ${z}_t$, and $\bm{\omega}_t$ denotes the standard Wiener process. The training of the diffusion model is achieved via denoising score matching:
\begin{align}\label{eqn:denosing}
    \mathcal{L} &= \mathbb{E}_{z_0 \sim p(z_0)} \mathbb{E}_{t \sim p(t)} \mathbb{E}_{\bm{\varepsilon}\sim \mathcal{N}(\bm{0}, I)} \Vert \bm{\epsilon}_{\theta}(z_t, t) - \bm{\varepsilon}\Vert_2^2, \nonumber \\
    &\text{where } z_t = z_0 + \sigma(t)\bm{\varepsilon},
\end{align}
where $\bm{\epsilon}_{\theta}$ is a denoising function learned via a neural network. In essence, it is an objective function aiming to predict the noise $\varepsilon$.  Samples are drawn by taking data from the base probability distribution provided in the latent space $p(z_{T})$ and solving the stochastic differential equation provided in the reverse process for each time step.

One diffusion model was created for each of the 17 vaccines in the dataset. The models were trained only on training observations to avoid potential data leakage. To validate our model, we took 5 unique samples of synthetic data for each diffusion model.

\subsection{Metrics}

\paragraph{Metrics for Predictive Models}

To evaluate model performance for \textbf{RQ1 and RQ2}, we measured class-weighted Precision, Recall, and F1-Score. These metrics were selected to provide a balanced assessment of model performance while accounting for class imbalances in the dataset. Predictions generated as probabilities were converted to binary classifications using a 0.5 decision threshold. All metrics were computed using the implementations provided in Scikit-learn.

\paragraph{Metrics for Synthetic Data}

To evaluate the synthetic data applied to \textbf{RQ2}, we assessed the fidelity of the generated output using the same metrics employed in the original TabSyn framework. Column-wise distributions were compared using the Kolmogorov–Smirnov Test (KST) for numerical variables and Total Variation Distance (TVD) for categorical variables. Relationships between feature pairs were evaluated using the Pearson Correlation Coefficient for numerical column pairs and Contingency Similarity for categorical column pairs. Collectively, these metrics measure how well the synthetic data preserve both marginal distributions and inter-feature dependencies present in the real dataset. Since we took the complement of these values, higher scores indicate greater similarity between real and synthetic data. Greater detail of the calculations required for these metrics can be found in the supplementary material.

 \section{Results}

\subsection{Predictive Results}

Model performance was evaluated on 5 different iterations of synthetic datasets. Examples of the precision, recall, and F1 scores for the BCG, ROTA2, and MR1 vaccines are shown in Table \ref{tab:small_model_results}, while full results are shown in Table \ref{tab:full_model_results} within the appendix.
\renewcommand{\arraystretch}{1.2}
\begin{table*}[t]
\begin{tabular}{llllllll}
\toprule
                 &                         & \multicolumn{3}{c}{\textbf{Logistic Regression}}                                    & \multicolumn{3}{c}{\textbf{XGBoost}}                            \\
\textbf{Vaccine} & \textbf{Training Split} & \multicolumn{1}{r}{\textbf{F1}}         & \textbf{Precision}  & \textbf{Recall}     & \textbf{F1}         & \textbf{Precision}  & \textbf{Recall}     \\
\midrule
BCG              & Real                    & 0.8348                                  & 0.9203              & 0.7921              & 0.8853              & 0.9239              & 0.863               \\
BCG              & Synthetic               & 0.8334 $\pm$ 0.0193                     & 0.9198 $\pm$ 0.0012 & 0.7909 $\pm$ 0.0245 & 0.8866 $\pm$ 0.0065 & 0.9173 $\pm$ 0.0015 & 0.867 $\pm$ 0.0102  \\
BCG              & Real + Synthetic        & 0.8234 $\pm$ 0.0089                     & 0.9209 $\pm$ 0.0011 & 0.7776 $\pm$ 0.0114 & 0.8807 $\pm$ 0.0051 & 0.9255 $\pm$ 0.011  & 0.857 $\pm$ 0.0073  \\
ROTA2& Real                    & 0.746                                   & 0.7485              & 0.7484              & 0.7501              & 0.7505              & 0.7513              \\
ROTA2& Synthetic               & 0.745 $\pm$ 0.006                       & 0.7482 $\pm$ 0.0059 & 0.7478 $\pm$ 0.0059 & 0.7391 $\pm$ 0.0067 & 0.7401 $\pm$ 0.0068 & 0.7412 $\pm$ 0.0067 \\
ROTA2& Real + Synthetic        & 0.7452 $\pm$ 0.0049                     & 0.7476 $\pm$ 0.005  & 0.7477 $\pm$ 0.005  & 0.7518 $\pm$ 0.0057 & 0.7524 $\pm$ 0.0059 & 0.7534 $\pm$ 0.0058 \\
MR1              & Real                    & 0.7053              & 0.7173              & 0.7046              & 0.7285              & 0.7362              & 0.7274              \\
MR1              & Synthetic               & 0.6996 $\pm$ 0.0069 & 0.7113 $\pm$ 0.0059 & 0.6993 $\pm$ 0.0074 & 0.7046 $\pm$ 0.0111 & 0.7158 $\pm$ 0.0098 & 0.7018 $\pm$ 0.0117 \\
MR1              & Real + Synthetic        & 0.7028 $\pm$ 0.0043 & 0.7139 $\pm$ 0.0044 & 0.7025 $\pm$ 0.0048 & 0.7201 $\pm$ 0.0039 & 0.7314 $\pm$ 0.0025 & 0.7184 $\pm$ 0.0045
\\
\bottomrule
\end{tabular}
\caption{Comparison of ML Model Performance (Mean ± Standard Deviation) across Iterations of Synthetic Data.}
\label{tab:small_model_results}
\end{table*}

The individual prediction models demonstrate strong performance for early vaccines, such as BCG, DPT1, and PCV1, with F1-scores exceeding 80\% in some cases. However, the models exhibit a decline in predictive capability for later vaccines in the cycle, such as MR and Vitamin A. This discrepancy may be attributed to a smaller dataset for later vaccines, as fewer children reach these stages of the vaccination schedule, and to increased variability in vaccination timing and compliance. Additionally, there are stronger class imbalances for earlier immunizations, leading to the models successfully predicting the majority class at an increased rate. Augmentation of synthetic data for late-schedule vaccines occasionally led to modest increases in model performance, while model performance sufficiently reproduced baseline results when replacing real with synthetic data for BCG.

These findings highlight the strengths of machine learning in predicting vaccination coverage for initial vaccines while pointing to the need for further refinement to address challenges associated with predicting later-stage vaccinations. By incorporating additional features or exploring advanced modeling techniques, future research can aim to improve accuracy across the entire vaccination cycle.

\subsection{Synthetic Data Distribution}
To validate the quality of our synthetic data, we present a series of low-order statistics for the BCG, DPT3, and MR1 vaccines. Since the TabSyn model for each synthetic dataset was trained on the same predictors and predictor values, the quality of the synthetic data is effectively independent of the different vaccines. Thus, the column-wise density and pairwise correlation scores for BCG, DPT3, and MR1 indicate the performance of the synthetic data across all vaccines. Scores closer to 1.0 indicate stronger similarity between the real and synthetic distributions.

\subsubsection{Column-Wise Density Estimation}

\begin{table*}[t]
\centering
\renewcommand{\arraystretch}{1.2}
\begin{tabular}{lcccl}
\toprule
\textbf{Column} 
& \textbf{BCG} 
& \textbf{DPT3} 
& \textbf{MR1}  &\textbf{All Vaccines}\\
\midrule
clinic          & 0.9866 $\pm$ 0.0026 & 0.9854 $\pm$ 0.0037 & 0.9898 $\pm$ 0.0045  &0.9888 $\pm$ 0.0044\\
age             & 0.9607 $\pm$ 0.0035 & 0.9716 $\pm$ 0.0050 & 0.9610 $\pm$ 0.0053  &0.9628 $\pm$ 0.0097\\
gender          & 0.9919 $\pm$ 0.0042 & 0.9010 $\pm$ 0.0067 & 0.9536 $\pm$ 0.0132  &0.9833 $\pm$ 0.0236\\
village         & 0.9457 $\pm$ 0.0035 & 0.9428 $\pm$ 0.0043 & 0.9637 $\pm$ 0.0031  &0.9529 $\pm$ 0.0086\\
first visit day & 0.9876 $\pm$ 0.0036 & 0.9797 $\pm$ 0.0034 & 0.9866 $\pm$ 0.0043  &0.9846 $\pm$ 0.0046\\
vax             & 0.9975 $\pm$ 0.0027 & 0.9971 $\pm$ 0.0024 & 0.9945 $\pm$ 0.0047  &0.9953 $\pm$ 0.0034\\
\bottomrule
\end{tabular}
\caption{Comparison of Column-Wise Density (Mean ± Standard Deviation) across Vaccines.}
\label{tab:merged_shape_scores}
\end{table*}
As evidenced by Table \ref{tab:merged_shape_scores}, the column-wise density scores indicate that the synthetic data distribution for each column successfully mimics the distribution of each variable. Categorical variables and the first visit day numeric variable had near-perfect similarity scores, while the age variable exhibited lower scores. 

\subsubsection{Pair-Wise Column Correlation}
The pair-wise column correlation scores for BCG, DPT3, and MR1 are shown in Table \ref{tab:column_pair_scores}. Similar to the Column-Wise Density estimation, scores closer to 1.0 indicate better inter-column correlation matching.

\begin{table*}[t]
\centering
\begin{tabular}{lcccc}
\toprule
\textbf{Column Pairs}& \textbf{BCG}& \textbf{DPT3}& \textbf{MR1}& \textbf{All Vaccines}\\
\midrule
clinic/age & 0.9347 $\pm$ 0.0045 & 0.9140 $\pm$ 0.0068 & 0.9195 $\pm$ 0.0050 & 0.9244 $\pm$ 0.0167 \\
clinic/gender & 0.9826 $\pm$ 0.0024 & 0.9003 $\pm$ 0.0064 & 0.9523 $\pm$ 0.0114 & 0.9732 $\pm$ 0.0210 \\
clinic/village & 0.9246 $\pm$ 0.0021 & 0.9274 $\pm$ 0.0049 & 0.9531 $\pm$ 0.0036 & 0.9392 $\pm$ 0.0143 \\
clinic/first visit day& 0.9803 $\pm$ 0.0018 & 0.9777 $\pm$ 0.0029 & 0.9820 $\pm$ 0.0056 & 0.9819 $\pm$ 0.0043 \\
clinic/vax & 0.9835 $\pm$ 0.0022 & 0.9684 $\pm$ 0.0058 & 0.9812 $\pm$ 0.0032 & 0.9803 $\pm$ 0.0060 \\
age/gender & 0.9574 $\pm$ 0.0034 & 0.8989 $\pm$ 0.0049 & 0.9423 $\pm$ 0.0054 & 0.9532 $\pm$ 0.0165 \\
age/village & 0.8550 $\pm$ 0.0027 & 0.8474 $\pm$ 0.0049 & 0.8528 $\pm$ 0.0044 & 0.8525 $\pm$ 0.0163 \\
age/first visit day& 0.9527 $\pm$ 0.0038 & 0.9626 $\pm$ 0.0058 & 0.9521 $\pm$ 0.0051 & 0.9546 $\pm$ 0.0096 \\
age/vax & 0.9563 $\pm$ 0.0028 & 0.9498 $\pm$ 0.0049 & 0.9413 $\pm$ 0.0050 & 0.9480 $\pm$ 0.0115 \\
gender/village & 0.9329 $\pm$ 0.0020 & 0.8847 $\pm$ 0.0092 & 0.9210 $\pm$ 0.0070 & 0.9321 $\pm$ 0.0153 \\
gender/first visit day& 0.9872 $\pm$ 0.0038 & 0.8974 $\pm$ 0.0064 & 0.9476 $\pm$ 0.0121 & 0.9781 $\pm$ 0.0233 \\
gender/vax & 0.9906 $\pm$ 0.0055 & 0.9010 $\pm$ 0.0067 & 0.9536 $\pm$ 0.0132 & 0.9805 $\pm$ 0.0228 \\
village/first visit day& 0.9340 $\pm$ 0.0031 & 0.9306 $\pm$ 0.0030 & 0.9528 $\pm$ 0.0023 & 0.9416 $\pm$ 0.0085 \\
village/vax & 0.9377 $\pm$ 0.0030 & 0.9271 $\pm$ 0.0048 & 0.9445 $\pm$ 0.0044 & 0.9380 $\pm$ 0.0077 \\
first visit day/vax& 0.9917 $\pm$ 0.0026 & 0.9901 $\pm$ 0.0025 & 0.9864 $\pm$ 0.0040 & 0.9891 $\pm$ 0.0035 \\
\bottomrule
\end{tabular}
\caption{Comparison of Pair-Wise Column Correlation Scores (Mean ± Standard Deviation) across Vaccines.}
\label{tab:column_pair_scores}
\end{table*}

Similarly to the column density estimations, the synthetic data accurately captures how correlated each variable is with one another, with the exception of variable pairs including age. The lower scores for age are likely attributed to the continuous nature and skewed distribution of the variable. TabSyn capably captured not only marginal distributions but relationships between predictors as well.

\subsubsection{Feature Importance Analysis}
After running the previous models, we analyzed feature importance from XGBoost models that were trained entirely on real data and entirely on the first sample of synthetic data. There are two reasons for performing this analysis: We wanted to identify the most influential features on vaccination coverage and wanted to ensure that the most notable features in the original models were also the most notable in the synthetic models. Importance values indicate the magnitude of influence each factor has on the likelihood of a child receiving a vaccine. 

\begin{figure}[htbp]
    \centering
    \includegraphics[width=\columnwidth]{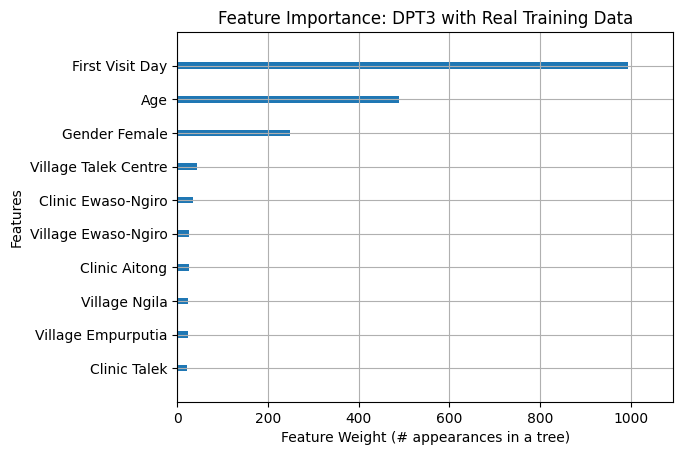}
    \caption{Feature Importance Analysis: DPT3 Real Data}
    \label{fig:feature_importance_real}
\end{figure}

\begin{figure}[htbp]
    \centering
    \includegraphics[width=\columnwidth]{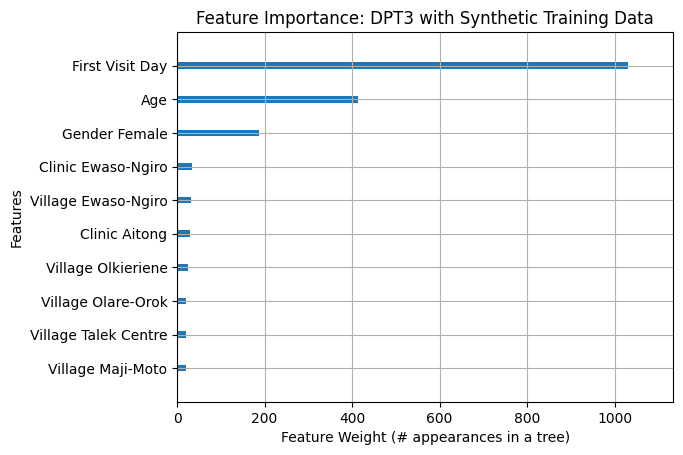}
    \caption{Feature Importance Analysis: DPT3 Synthetic Data}
    \label{fig:feature_importance_syn}
\end{figure}

As shown in Figure \ref{fig:feature_importance_real}, key features such as the day of first visit, current age of the child, and gender of the child emerged as significant predictors. This finding aligns with insights from the descriptive analysis, where contextual and demographic factors were identified as critical determinants of vaccination coverage. Specifically:
\begin{itemize}
    \item \textbf{First Visit Day:} Strongly influences vaccination likelihood, likely due to correlations between earlier visits and vaccine access.
    \item \textbf{Age:} The current age of a child seems to be influential and could account for differences in vaccine supply throughout the duration of the dataset.
    \item \textbf{Gender:} Although the overall gender distribution of vaccinated children is equal, gender remains a significant factor in individual-level predictions.
\end{itemize}

These results provide actionable insights, suggesting that early targeted interventions, specifically for girls, could significantly improve vaccination rates. Additionally, they emphasize the importance of accounting for demographic nuances in predictive modeling. In terms of overall performance, our results show how TabSyn can effectively capture the distribution of vaccination data in Narok, and how we can replace real data with the synthetically generated data without any significant loss in prediction ability. 

Our results also display how synthetic data strategies can be implemented into primary health care solutions. In many LMIC settings, such as CHP, which aren't fully technologically equipped for EHRs, this concept can be applied to assist in integrating digital clinical assistance tools into health systems where the burden of electronic data collection is typically too large a barrier to utilize them for interventions. The team of researchers within the Ignite Lab spent hundreds of hours in digitizing the records for the original dataset; it took on average 90 minutes to train the VAE and Diffusion models locally on a CPU, and only approximately 20 seconds to sample a synthetic dataset equal in size to the original. This illustrates both the scaling power as well as the replicability of this study within the context of LMIC settings via local training without extensive computational resources.


\section{Discussion}

This study highlights the critical role of data-driven approaches in addressing healthcare disparities in underserved communities. By leveraging machine learning techniques and integrating interdisciplinary real and synthetic data, we demonstrated the potential to predict childhood vaccination demand with high accuracy. Our findings underscore the importance of geographic, demographic, and cultural factors in shaping healthcare outcomes, particularly in remote regions like Narok County.

Our results are crucial for multiple reasons. First, our results display a proof of concept for how synthetic data augmentation strategies can be implemented into primary health care solutions. The ability to artificially generate new digital data quickly allows for more efficient scalability in community health clinics like CHP, which are not currently equipped with Electronic Health Records (EHRs) and digital recordkeeping. This concept can be applied to assist in integrating digital clinical assistance tools into health systems where the burden of electronic data collection is typically too large a barrier to utilize them for interventions. While it's still important to encourage the use of EHRs where possible to improve the timeliness and accuracy of data, synthetic data can be used to help achieve accurate insights quickly. 

However, with these results, there are two crucial limitations to the study. First, the quantity of data may be influenced by the catchment population, including individuals who did not attend CHP clinics, potentially leading to the underrepresentation of certain demographics and vaccines in the dataset. Additionally, errors in data entry or misclassification could affect variable definitions and analyses, introducing potential bias or measurement inaccuracies. While this was offset in part by removing observations that had missing data or obviously incorrect features, the possibility for unseen errors remains. Together, the limitations in the data may affect the generalizability of our findings.

Moving forward, our collaboration with Community Health Partners (CHP) and the refinement of predictive models will ensure that our solutions remain both actionable and culturally sensitive. By focusing on improving model performance, expanding data integration, and developing practical tools for healthcare providers, we aim to contribute to more equitable and efficient healthcare resource allocation.

This work represents a step toward bridging the gap between data science and global health equity. Through continued innovation and collaboration, we hope to advance the reach and impact of vaccination programs, ultimately improving health outcomes for vulnerable populations in Narok and similar resource-constrained settings.

\section{Acknowledgement}

We are grateful to the community health partners in Narok for their collaboration, insights, and sustained engagement throughout this study.  We also acknowledge the Applied Research and Innovation Initiative (ARII) at William \& Mary for providing the financial support that made this work possible, as well as the Global Research Institute at William \& Mary for funding the Summer Fellows Program. Finally, we thank all students who contributed to the digitization study between 2017 and 2023.

\section{Data Availability}
The datasets generated or analyzed during this study are not publicly available to protect patient confidentiality and prevent harm resulting from misuse of sensitive health information but are available from the corresponding author on reasonable request.

\section{Authors' Contributions}
Conceptualization: JB, HC, CBD, JNO
Data Curation: JB, Y Liu
Formal Analysis: JB, Y Liu
Funding Acquisition: HC, CBD, JNO
Methodology: JB, Y Li, HC
Project Administration: JB, HC
Resources: JS, RK
Software: JB
Supervision: Y Li, HC, CBD
Visualization: JB, Y Liu
Writing (original draft): JB, JNO, CBD, HC
Writing (review and editing): Y Li, JB

\bibliographystyle{unsrt}
\bibliography{kenya}
\clearpage

\appendix

\section{Full Modeling Results}\label{appendix:full_modeling}

\begin{table}[h]
\begin{tabular}{llllllll}
\toprule
                 &                         & \multicolumn{3}{c}{\textbf{Logistic Regression}}                                    & \multicolumn{3}{c}{\textbf{XGBoost}}                            \\
\textbf{Vaccine} & \textbf{Training Split} & \multicolumn{1}{r}{\textbf{F1}}         & \textbf{Precision}  & \textbf{Recall}     & \textbf{F1}         & \textbf{Precision}  & \textbf{Recall}     \\
\midrule
MR1              & Real                    & 0.7053                                  & 0.7173              & 0.7046              & 0.7285              & 0.7362              & 0.7274              \\
MR1              & Synthetic               & 0.6996 $\pm$ 0.0069                     & 0.7113 $\pm$ 0.0059 & 0.6993 $\pm$ 0.0074 & 0.7046 $\pm$ 0.0111 & 0.7158 $\pm$ 0.0098 & 0.7018 $\pm$ 0.0117 \\
MR1              & Real + Synthetic        & 0.7028 $\pm$ 0.0043                     & 0.7139 $\pm$ 0.0044 & 0.7025 $\pm$ 0.0048 & 0.7201 $\pm$ 0.0039 & 0.7314 $\pm$ 0.0025 & 0.7184 $\pm$ 0.0045 \\
MR2              & Real                    & 0.788                                   & 0.8438              & 0.7849              & 0.8172              & 0.8407              & 0.8145              \\
MR2              & Synthetic               & 0.7788 $\pm$ 0.0083                     & 0.8415 $\pm$ 0.0058 & 0.7737 $\pm$ 0.0088 & 0.8097 $\pm$ 0.0054 & 0.8376 $\pm$ 0.0063 & 0.806 $\pm$ 0.0065  \\
MR2              & Real + Synthetic        & 0.7854 $\pm$ 0.0055                     & 0.8433 $\pm$ 0.0039 & 0.783 $\pm$ 0.005   & 0.8122 $\pm$ 0.0049 & 0.8432 $\pm$ 0.0058 & 0.8075 $\pm$ 0.0051 \\
BCG              & Real                    & 0.8348                                  & 0.9203              & 0.7921              & 0.8853              & 0.9239              & 0.863               \\
BCG              & Synthetic               & \multicolumn{1}{r}{0.8334 $\pm$ 0.0193} & 0.9198 $\pm$ 0.0012 & 0.7909 $\pm$ 0.0245 & 0.8866 $\pm$ 0.0065 & 0.9173 $\pm$ 0.0015 & 0.867 $\pm$ 0.0102  \\
BCG              & Real + Synthetic        & \multicolumn{1}{r}{0.8234 $\pm$ 0.0089} & 0.9209 $\pm$ 0.0011 & 0.7776 $\pm$ 0.0114 & 0.8807 $\pm$ 0.0051 & 0.9255 $\pm$ 0.011  & 0.857 $\pm$ 0.0073  \\
DPT1             & Real                    & 0.7522                                  & 0.8111              & 0.7581              & 0.7723              & 0.8009              & 0.7686              \\
DPT1             & Synthetic               & 0.752 $\pm$ 0.0049                      & 0.7959 $\pm$ 0.0335 & 0.7623 $\pm$ 0.0051 & 0.7692 $\pm$ 0.0076 & 0.7928 $\pm$ 0.0057 & 0.7723 $\pm$ 0.0096 \\
DPT1             & Real + Synthetic        & 0.7562 $\pm$ 0.0046                     & 0.8193 $\pm$ 0.0185 & 0.7664 $\pm$ 0.0048 & 0.7669 $\pm$ 0.0047 & 0.798 $\pm$ 0.0049  & 0.7659 $\pm$ 0.0056 \\
DPT2             & Real                    & 0.6979                                  & 0.7221              & 0.7097              & 0.7106              & 0.7244              & 0.7144              \\
DPT2             & Synthetic               & 0.6875 $\pm$ 0.0074                     & 0.7127 $\pm$ 0.0058 & 0.7084 $\pm$ 0.0055 & 0.6957 $\pm$ 0.0118 & 0.7094 $\pm$ 0.0104 & 0.7048 $\pm$ 0.0104 \\
DPT2             & Real + Synthetic        & 0.6963 $\pm$ 0.0052                     & 0.7194 $\pm$ 0.0034 & 0.7112 $\pm$ 0.0037 & 0.7005 $\pm$ 0.0062 & 0.7162 $\pm$ 0.0063 & 0.7077 $\pm$ 0.0056 \\
DPT3             & Real                    & 0.6835                                  & 0.6879              & 0.6858              & 0.6876              & 0.6903              & 0.6898              \\
DPT3             & Synthetic               & 0.6713 $\pm$ 0.0044                     & 0.6769 $\pm$ 0.0048 & 0.6772 $\pm$ 0.0044 & 0.6723 $\pm$ 0.0123 & 0.6753 $\pm$ 0.0119 & 0.6779 $\pm$ 0.013  \\
DPT3             & Real + Synthetic        & 0.6821 $\pm$ 0.0029                     & 0.6856 $\pm$ 0.0034 & 0.686 $\pm$ 0.0031  & 0.6816 $\pm$ 0.0105 & 0.6846 $\pm$ 0.0103 & 0.685 $\pm$ 0.0107  \\
IPV              & Real                    & 0.6987                                  & 0.7023              & 0.7007              & 0.7092              & 0.7105              & 0.7101              \\
IPV              & Synthetic               & 0.6968 $\pm$ 0.0076                     & 0.7003 $\pm$ 0.0084 & 0.6988 $\pm$ 0.0079 & 0.7054 $\pm$ 0.006  & 0.709 $\pm$ 0.006   & 0.7074 $\pm$ 0.0058 \\
IPV              & Real + Synthetic        & 0.6987 $\pm$ 0.0037                     & 0.7023 $\pm$ 0.0038 & 0.7006 $\pm$ 0.0037 & 0.7139 $\pm$ 0.0065 & 0.7167 $\pm$ 0.006  & 0.7155 $\pm$ 0.0062 \\
OPV1             & Real                    & 0.7571                                  & 0.7886              & 0.7654              & 0.7698              & 0.7811              & 0.7672              \\
OPV1             & Synthetic               & 0.7401 $\pm$ 0.0099                     & 0.7779 $\pm$ 0.0079 & 0.7512 $\pm$ 0.0083 & 0.7546 $\pm$ 0.0093 & 0.7742 $\pm$ 0.0096 & 0.7557 $\pm$ 0.01   \\
OPV1             & Real + Synthetic        & 0.7498 $\pm$ 0.0064                     & 0.7838 $\pm$ 0.0046 & 0.7597 $\pm$ 0.0039 & 0.7653 $\pm$ 0.0101 & 0.7822 $\pm$ 0.01   & 0.7639 $\pm$ 0.0115 \\
OPV2             & Real                    & 0.7074                                  & 0.7182              & 0.7079              & 0.7235              & 0.7281              & 0.726               \\
OPV2             & Synthetic               & 0.7139 $\pm$ 0.0081                     & 0.7225 $\pm$ 0.0077 & 0.7157 $\pm$ 0.0071 & 0.7066 $\pm$ 0.0092 & 0.7115 $\pm$ 0.0094 & 0.7094 $\pm$ 0.0095 \\
OPV2             & Real + Synthetic        & 0.7167 $\pm$ 0.0055                     & 0.7255 $\pm$ 0.0054 & 0.7173 $\pm$ 0.0054 & 0.7152 $\pm$ 0.006  & 0.7198 $\pm$ 0.0053 & 0.7171 $\pm$ 0.0063 \\
OPV3             & Real                    & 0.701                                   & 0.7021              & 0.7035              & 0.6901              & 0.6908              & 0.6913              \\
OPV3             & Synthetic               & 0.699 $\pm$ 0.005                       & 0.7011 $\pm$ 0.0054 & 0.7025 $\pm$ 0.0054 & 0.6927 $\pm$ 0.0093 & 0.6936 $\pm$ 0.0094 & 0.6954 $\pm$ 0.009  \\
OPV3             & Real + Synthetic        & 0.7009 $\pm$ 0.0044                     & 0.7026 $\pm$ 0.0045 & 0.7038 $\pm$ 0.0046 & 0.6993 $\pm$ 0.0073 & 0.7004 $\pm$ 0.0073 & 0.7011 $\pm$ 0.0073 \\
PCV1             & Real                    & 0.7468                                  & 0.8179              & 0.7538              & 0.7739              & 0.8024              & 0.773               \\
PCV1             & Synthetic               & 0.7414 $\pm$ 0.0027                     & 0.7863 $\pm$ 0.0325 & 0.7521 $\pm$ 0.0027 & 0.7496 $\pm$ 0.0074 & 0.78 $\pm$ 0.0089   & 0.7496 $\pm$ 0.0072 \\
PCV1             & Real + Synthetic        & 0.7479 $\pm$ 0.0026                     & 0.7909 $\pm$ 0.0176 & 0.759 $\pm$ 0.0023  & 0.761 $\pm$ 0.0067  & 0.7939 $\pm$ 0.0059 & 0.7605 $\pm$ 0.0066 \\
PCV2             & Real                    & 0.6988                                  & 0.7226              & 0.7115              & 0.7103              & 0.7199              & 0.7162              \\
PCV2             & Synthetic               & 0.685 $\pm$ 0.0104                      & 0.7086 $\pm$ 0.0088 & 0.7071 $\pm$ 0.0076 & 0.6882 $\pm$ 0.0098 & 0.7005 $\pm$ 0.0101 & 0.6971 $\pm$ 0.0083 \\
PCV2             & Real + Synthetic        & 0.6927 $\pm$ 0.0064                     & 0.7167 $\pm$ 0.0059 & 0.7082 $\pm$ 0.0049 & 0.7013 $\pm$ 0.0057 & 0.7166 $\pm$ 0.0051 & 0.7082 $\pm$ 0.0052 \\
PCV3             & Real                    & 0.688                                   & 0.6915              & 0.6916              & 0.692               & 0.6936              & 0.6941              \\
PCV3             & Synthetic               & 0.6812 $\pm$ 0.0031                     & 0.6854 $\pm$ 0.0031 & 0.6857 $\pm$ 0.0023 & 0.6724 $\pm$ 0.0105 & 0.6753 $\pm$ 0.0112 & 0.6762 $\pm$ 0.0107 \\
PCV3             & Real + Synthetic        & 0.6887 $\pm$ 0.0047                     & 0.6922 $\pm$ 0.0043 & 0.6925 $\pm$ 0.0046 & 0.6888 $\pm$ 0.0087 & 0.6915 $\pm$ 0.0084 & 0.6923 $\pm$ 0.0087 \\
\bottomrule
\end{tabular}
\textit{Continues on next page.}
\end{table}

\begin{table*}[h]
\begin{tabular}{llllllll}
\toprule
                 &                         & \multicolumn{3}{c}{\textbf{Logistic Regression}}                                    & \multicolumn{3}{c}{\textbf{XGBoost}}                            \\
\textbf{Vaccine} & \textbf{Training Split} & \multicolumn{1}{r}{\textbf{F1}}         & \textbf{Precision}  & \textbf{Recall}     & \textbf{F1}         & \textbf{Precision}  & \textbf{Recall}     \\
\midrule
POLIO            & Real                    & 0.7239                                  & 0.7243              & 0.7242              & 0.7632              & 0.764               & 0.7636              \\
POLIO            & Synthetic               & 0.7188 $\pm$ 0.0069                     & 0.719 $\pm$ 0.0068  & 0.7189 $\pm$ 0.0068 & 0.7535 $\pm$ 0.0109 & 0.754 $\pm$ 0.011   & 0.7537 $\pm$ 0.0109 \\
POLIO            & Real + Synthetic        & 0.7212 $\pm$ 0.0048                     & 0.7214 $\pm$ 0.0048 & 0.7213 $\pm$ 0.0048 & 0.7677 $\pm$ 0.0091 & 0.7684 $\pm$ 0.0092 & 0.768 $\pm$ 0.0091  \\
ROTA1            & Real                    & 0.7907                                  & 0.7961              & 0.7889              & 0.7842              & 0.7881              & 0.7827              \\
ROTA1            & Synthetic               & 0.7875 $\pm$ 0.0036                     & 0.7932 $\pm$ 0.0029 & 0.7865 $\pm$ 0.0034 & 0.7793 $\pm$ 0.0085 & 0.7836 $\pm$ 0.008  & 0.779 $\pm$ 0.0082  \\
ROTA1            & Real + Synthetic        & 0.7907 $\pm$ 0.0045                     & 0.7958 $\pm$ 0.0036 & 0.7895 $\pm$ 0.0045 & 0.7883 $\pm$ 0.0017 & 0.7939 $\pm$ 0.0018 & 0.7868 $\pm$ 0.0017 \\
ROTA2            & Real                    & 0.746                                   & 0.7485              & 0.7484              & 0.7501              & 0.7505              & 0.7513              \\
ROTA2            & Synthetic               & 0.745 $\pm$ 0.006                       & 0.7482 $\pm$ 0.0059 & 0.7478 $\pm$ 0.0059 & 0.7391 $\pm$ 0.0067 & 0.7401 $\pm$ 0.0068 & 0.7412 $\pm$ 0.0067 \\
ROTA2            & Real + Synthetic        & 0.7452 $\pm$ 0.0049                     & 0.7476 $\pm$ 0.005  & 0.7477 $\pm$ 0.005  & 0.7518 $\pm$ 0.0057 & 0.7524 $\pm$ 0.0059 & 0.7534 $\pm$ 0.0058 \\
VITA             & Real                    & 0.7414                                  & 0.7464              & 0.7411              & 0.7431              & 0.7477              & 0.7422              \\
VITA             & Synthetic               & 0.7309 $\pm$ 0.0065                     & 0.7382 $\pm$ 0.0064 & 0.7309 $\pm$ 0.0067 & 0.7272 $\pm$ 0.0065 & 0.7329 $\pm$ 0.0046 & 0.7261 $\pm$ 0.0069 \\
VITA             & Real + Synthetic        & 0.7382 $\pm$ 0.0039                     & 0.745 $\pm$ 0.0044  & 0.7383 $\pm$ 0.004  & 0.7404 $\pm$ 0.0046 & 0.7467 $\pm$ 0.0045 & 0.7396 $\pm$ 0.0045 \\
\bottomrule
\end{tabular}
\caption{Full Machine Learning Results}
\label{tab:full_model_results}
\end{table*}
\end{document}

%% file: table2.tex
\begin{tabular}{c|c|c}
\toprule
\textbf{Variable}                            & \textbf{Categorical Value}            & \textbf{Overall}     \\
\hline
 n                          &             & 6912        \\
 \hline
 Age (yrs), mean (SD)       &             & 3.3 (2.1)   \\
 \hline
 First Visit Day, mean (SD) &             & 24.9 (55.2) \\
 \hline
 Gender, n (\%)              & Female      & 3396 (49.1) \\
                            & Male        & 3516 (50.9) \\
\hline
 Clinic, n (\%)              & Aitong      & 1523 (22.0) \\
                            & Ewaso-Ngiro & 3425 (49.6) \\
                            & Mara Rianta & 912 (13.2)  \\
                            & Talek       & 1052 (15.2) \\
\hline
\end{tabular}

%% file: table3.tex
\begin{tabular}{l c}
\toprule
\textbf{Vaccine} & \textbf{n = 6913, No / Yes (\%)} \\
\midrule
BCG  & 301 (4.4) / 6612 (95.6) \\
POLIO & 3310 (47.9) / 3603 (52.1) \\
OPV1 & 1476 (21.4) / 5437 (78.6) \\
OPV2 & 2177 (31.5) / 4736 (68.5) \\
OPV3 & 2894 (41.9) / 4019 (58.1) \\
IPV  & 3208 (46.4) / 3705 (53.6) \\
DPT1 & 1076 (15.6) / 5837 (84.4) \\
DPT2 & 1817 (26.3) / 5096 (73.7) \\
DPT3 & 2532 (36.6) / 4381 (63.4) \\
PCV1 & 1088 (15.7) / 5825 (84.3) \\
PCV2 & 1817 (26.3) / 5096 (73.7) \\
PCV3 & 2550 (36.9) / 4363 (63.1) \\
ROTA1 & 2103 (30.4) / 4810 (69.6) \\
ROTA2 & 2901 (42.0) / 4012 (58.0) \\
VITA  & 4222 (61.1) / 2691 (38.9) \\
MR1   & 4298 (62.2) / 2615 (37.8) \\
MR2   & 5963 (86.3) / 950 (13.7) \\
\bottomrule
\end{tabular}